\documentclass{article}

\usepackage[preprint]{neurips_2024}

\usepackage[utf8]{inputenc} 
\usepackage[T1]{fontenc}    
\usepackage{hyperref}       
\usepackage{url}            
\usepackage{booktabs}       
\usepackage{amsfonts}       
\usepackage{nicefrac}       
\usepackage{microtype}      
\usepackage{xcolor}         
\usepackage{graphics}
\usepackage[pdftex]{graphicx}

\hypersetup{hidelinks} 
\usepackage{multirow}
\usepackage{mathtools,amssymb}
\usepackage{enumitem}
\usepackage{float}
\usepackage{wrapfig}
\usepackage[vlined,ruled,linesnumbered]{algorithm2e}
\title{CLAQ: Pushing the Limits of Low-Bit Post-Training Quantization for LLMs}

\author{%
  Haoyu Wang\thanks{Equal Contribution. This work was done when Haoyu Wang and Bei Liu were interns at Meituan.}\\
  Shanghai Jiao Tong University \\
  Shanghai, China \\
  \texttt{fayuge@sjtu.edu.cn} \\
  \And
  Bei Liu$^*$\\
  Shanghai Jiao Tong University \\
  Shanghai, China \\
  \texttt{beiliu@sjtu.edu.cn} \\
  \And
  Hang Shao \\
  Shanghai Jiao Tong University \\
  Shanghai, China \\
  \texttt{hangshao99@sjtu.edu.cn} \\
  \And
  Bo Xiao \\
  Meituan \\
  Beijing, China \\
  \texttt{xiaobo09@meituan.com} \\
  \And
  Ke Zeng \\
  Meituan \\
  Beijing, China \\
  \texttt{zengke02@meituan.com} \\
  \And
  Guanglu Wan \\
  Meituan \\
  Beijing, China \\
  \texttt{wanguanglu@meituan.com} \\
  \And
  Yanmin Qian\thanks{Corresponding Author.} \\
  Shanghai Jiao Tong University \\
  Shanghai, China \\
  yanminqian@sjtu.edu.cn \\
}

\begin{document}

\maketitle

\begin{abstract}
Parameter quantization for Large Language Models (LLMs) has attracted increasing attentions recently in reducing memory costs and improving computational efficiency. Early approaches have been widely adopted. However, the existing methods suffer from poor performance in low-bit (such as 2 to 3 bits) scenarios. In this paper, we present a novel and effective \textbf{C}olumn-\textbf{L}evel \textbf{A}daptive weight \textbf{Q}uantization (CLAQ) framework by introducing three different types of adaptive strategies for LLM quantization. Firstly, a K-Means clustering based algorithm is proposed that allows dynamic generation of quantization centroids for each column of a parameter matrix. Secondly, we design an outlier-guided adaptive precision search strategy which can dynamically assign varying bit-widths to different columns. Finally, a dynamic outlier reservation scheme is developed to retain some parameters in their original float point precision, in trade off of boosted model performance. Experiments on various mainstream open source LLMs including LLaMA-1, LLaMA-2 and Yi demonstrate that our methods achieve the state-of-the-art results across different bit settings, especially in extremely low-bit scenarios. Code is available at \href{https://github.com/fayuge/CLAQ}{https://github.com/fayuge/CLAQ}.
\end{abstract}


\section{Introduction}
Recent advancements in Large Language Models (LLMs) have yielded impressive accomplishments across a variety of tasks. Building on the pioneering work of GPT-3 \cite{gpt3}, a series of influential LLMs such as OPT \cite{zhang2022opt}, BLOOM \cite{le2023bloom}, LLaMA \cite{touvron2023LLaMA}, Mixtral \cite{jiang2024mixtral} and Yi \cite{young2024yi} have consistently demonstrated that expanding model size predictably promotes modeling accuracy (\textit{a.k.a.} the scaling law \cite{henighan2020scaling}). The pursue of large (\textit{e.g.} hundreds of billions of parameters), and even larger language models is often considered as a promising avenue towards Artificial General Intelligence (AGI) \cite{bubeck2023sparks}.


Nevertheless, the significant computational and memory demands of LLMs pose unprecedented challenges. For example, deploying the GPT-3 model requires a staggering 350GB of memory merely to accommodate its parameters in half-precision format (FP16), with a minimum of five high-end A100-80G GPUs. Such immense memory demands, coupled with the enormous runtime communication overhead, hinder the integration of LLMs into practical applications. To tackle this problem, 8-bit Post-Training Quantization (PTQ) methods for both model weights and activations have been proposed \cite{xiao2023smoothquant, dettmers2022int8}, which have led to reduced model sizes in memory and negligible declines in accuracy. Unfortunately, when PTQ methods \cite{chee2024quip,lin2023awq,kim2023squeezellm} are attempted in quantizing the parameters into 4 bits or less, notable decreases of performance have been observed, making further reduction of model sizes a challenging task. Additionally, several quantization-aware training (QAT) approaches \cite{liu2023llmqat,dettmers2024qlora,li2023loftq} have been introduced to enhance the model performance. Nonetheless, these algorithms integrate a training phase within the quantization process, thus imposing substantial demands for computational resources.

In this paper, we introduce a novel and effective training-free quantization framework for LLMs, namely \textbf{C}olumn-\textbf{L}evel \textbf{A}daptive weight \textbf{Q}uantization (CLAQ), which utilizes three distinct column-level strategies to enhance the performance of low-bit (\text{i.e.}, 2-bit and 3-bit) quantization. Specifically, we initially apply a K-Means clustering based quantization that enables the adaptive generation of quantization centroids for each column of a weight matrix. Then, a simple yet powerful quantization sensitivity metric (Outlier Order) is described, which aims to identify parameter columns that are highly sensitive to quantization. Leveraging this metric, we further proposed two strategies to enhance quantization performance, including Column-Level Adaptive Precision (AP) quantization and Column-Level Adaptive Outlier Reservation (OR). Both schemes can significantly boost the performance of the low-bit quantized LLMs with only a slight  increase in memory footprint. Notably, our fusion model (AP+OR) yields the best results in extremely low-bit scenarios. Extensive experiments on mainstream public LLMs and benchmarks demonstrate that our method achieves the state-of-the-art performance across various quantization bit-widths. 

\begin{figure}
  \centering
  \includegraphics[width=0.8\linewidth]{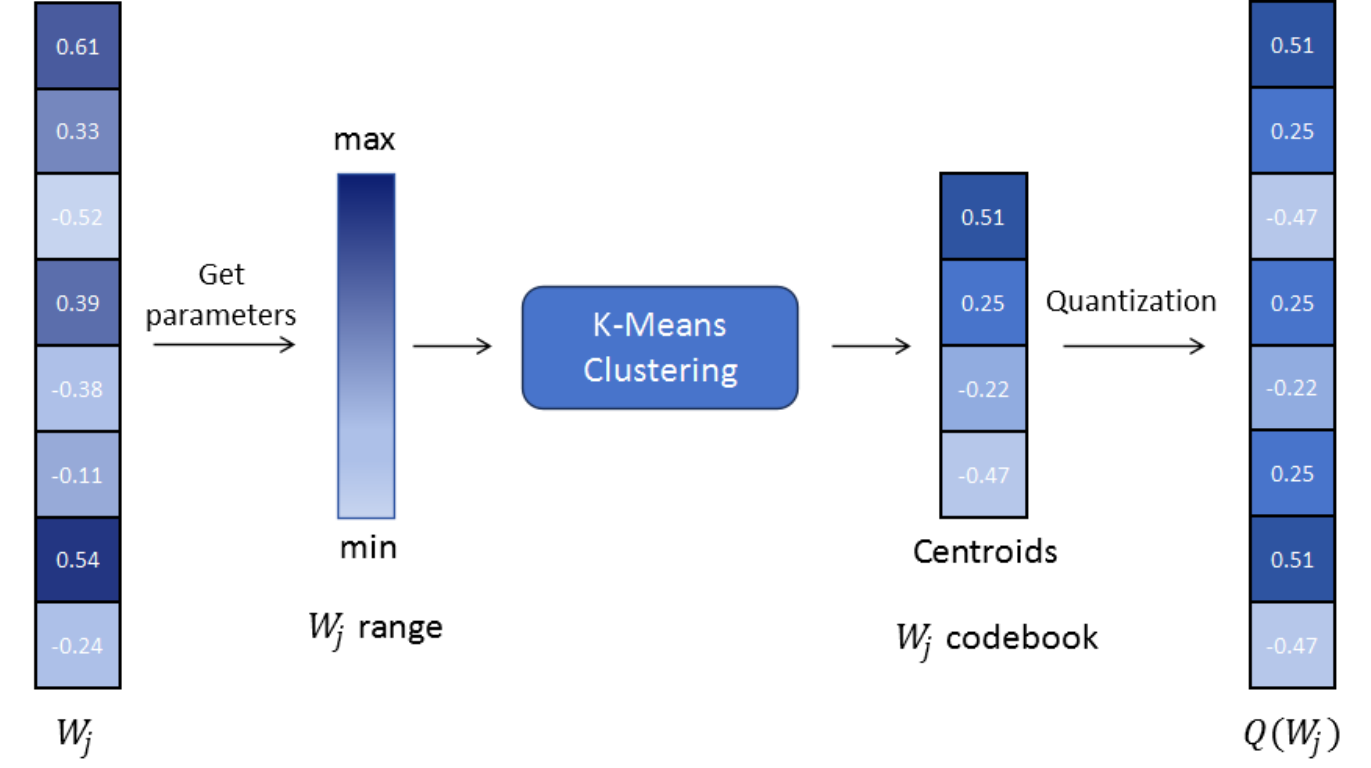}
  \caption{The K-Means clustering based quantization. Elements in weight matrix column are input of K-Means clustering and the quantization centroids are derived as the output of clustering algorithm. Then the pre-trained weights are quantized to the nearest K-Means class center.}
  \label{fig:1}
\end{figure}

The main contributions of our work can be summarized as follows:
\begin{enumerate}[leftmargin=*]
\item We introduce a novel centroid selection strategy for quantization: K-Means based weight quantization. The architecture of K-Means based quantization is illustrated in Figure \ref{fig:1}. Compared to existing methods, K-Means based quantization excels in more accurately approximating the parameter distribution of pre-trained models and dynamically generating quantization centroids. Our proposed quantization scheme surpasses all current 3-bit and 4-bit PTQ methods under equivalent or larger quantization group sizes.

\item We present a column-wise outlier ratio based quantization sensitivity metric, namely Outlier Order, to assess the importance of parameters in LLMs. Based on this metric, Adaptive Precision (AP) quantization and Outlier Reservation (OR) are proposed to allocate higher budget to columns identified as more sensitive to quantization errors. Notably, this sensitivity metric emerges as an extremely efficient guiding principle, which can be calculated once easily and directly applied to both AP and OR.

\item We implement Adaptive Precision quantization with the guidance of Outlier Order. Different levels of precision are adaptively assigned within each weight matrix. By allocating higher quantization budget to columns with a greater concentration of outliers, we achieve a substantial reduction in quantization errors. Furthermore, adaptive precision quantization allows arbitrary equivalent bit-width from 2-bit to 4-bit, facilitating the deployment of LLMs across a variety of end-devices with different memory constraints.

\item We develop a novel Outliers Reservation (OR) strategy to further enhance the performance with only a slight  increase in memory footprint. Based on Outlier Order, OR involves reserving FP16 precision outliers for columns with a higher concentration of outliers to preserve model performance, whereas columns with a lower outlier ratio are allocated less reservation budget. Moreover, the OR strategy can be combined with AP quantization, which exhibits the best results in low-bit quantization.

\end{enumerate}

\section{LLM Quantization}

\paragraph{Post-Training Quantization}
Post-training quantization methods \cite{wang2020towards,hubara2021accurate,li2021brecq,nagel2020up} quantize LLMs without fine-tuning the parameters, while compensating the quantization error through specially designed techniques. Previous work \cite{xiao2023smoothquant,yao2022zeroquant,dettmers2022int8,park2022qumm,Dettmers2022tim} have utilized the vanilla Round To Nearest quantization. Based on the OBS approach \cite{frantar2022OBQ}, GPTQ \cite{frantar2022gptq} adjusted the remaining weights while quantizing, offering an effective PTQ paradigm. A series of work have \cite{lin2023awq, xiao2023smoothquant, lee2024owq, dettmers2023spqr, yin2023owl, heo2023rethinking, wei2022outliersupp} noticed the significance of outlier treatment in LLM quantization. By preserving outliers, the performance of quantized models can be further improved. PTQ approach alleviates the computational burden associated with extensive re-training, thus facilitating the efficient deployment of LLMs in resource-constrained environments.

\paragraph{Quantization-Aware Training}
Quantization-Aware Training (QAT) incorporates a training phase within the quantization workflow. Owing to the model's ability to adapt to the constraints imposed by quantization, QAT typically leads to superior model performance. \cite{li2023loftq,dettmers2024qlora,xu2023qa} integrated quantization with the basis of LoRA \cite{hu2021lora}. \cite{xu2024onebit} performed extreme quantization with 1-bit parameters. In \cite{shao2023omniquant}, instead of training the quantized parameters, a learnable equivalent transformation is trained for quantization. By learning the quantization error, QAT can mitigate the information loss brought by quantization schemes. Nonetheless, the advantage of QAT comes with the inclusion of specialized training phases, which not only requires additional time but also increases the computational overhead and complexity of the quantization task. In practical applications, it may also couple with model fine-tuning processes that complicates the workflow. Therefore, in this work, we focus on PTQ approach first, and leave the study of QAT approach as a future work.

\section{CLAQ}


\subsection{K-Means Based Quantization}
In general, prior quantization techniques on LLMs have typically adopted uniform quantization levels (\textit{i.e.}, quantization centroids are equally spaced), such as SmoothQuant \cite{xiao2023smoothquant} and GPTQ \cite{frantar2022gptq}. Alternative methods adopt strategies that sample quantization centroids from a normal distribution, as demonstrated by QLoRA \cite{dettmers2024qlora}. Nonetheless, all of the above approaches fail to accurately capture the true distribution of a group of model parameters, given the heterogeneous nature of parameter magnitudes and distributions across various network layers. 


To tackle this issue, we first introduce a straightforward yet effective strategy: employing K-Means clustering for the selection of quantization centroids (Figure \ref{fig:1}). The clustering samples (\textit{i.e.}, quantization groups) are the columns of a parameter matrix from either self-attention or MLP structures in the LLMs.

Specifically, when quantizing an parameter matrix $W$ with shape $i \times j$  
, the formation of the quantization codebook $C_{W_j}$ for its $j$-th column is derived as follows:

\begin{equation}
C_{W_j} = KMeans(W_j)
\end{equation}

Where $KMeans$ denotes the clustering operation of K-Means algorithm, where the number of cluster centroids $K$ is determined by the target precision of the quantization, \textit{i.e.}, $2^{bit}$ centroids in total. For instance, in 4-bit quantization, the codebook size is $2^4=16$, implying the need of 16 distinct quantization target values. Once these cluster centroids are obtained, they are employed as the discrete quantization target values. Each parameter is quantized to its nearest centroid in the codebook, and is replaced by the codebook index in storage. For de-quantization, full-precision parameters can be retrieved from the codebook according to these indices. Specifically, for any parameter $W_{ij}$ in the model, the quantization operation $Q$ can be formalized as:

\begin{equation}
q\in \left \{ 0,1,2,\cdot \cdot \cdot ,2^{bit}-1 \right \} ,   \quad     Q(W_{ij}) = \mathop{\arg\min}_{q} |C_{W_j}(q)-W_{ij}|
\end{equation}


Inspired by \cite{frantar2022gptq}, our approach quantizes the parameter matrix $W$ per column through clustering and quantization. We adopt the same approach as GPTQ \cite{frantar2022gptq} for updating the remaining parameters. The advantage of K-Means based quantization is three-fold: Firstly, the centroids generated by K-Means clustering adaptively conform to the distribution of the LLM's parameters. The better matching with the underlying parameter distribution can minimize information loss and mitigate quantization errors. Secondly, our K-Means quantization reduces storage requirements by storing a single vector of K-Means centroids as codebook per column, yet not losing the performance. Previous approaches have employed smaller quantization groups to boost low-bit quantization performance. However, smaller groups incur additional overhead of codebook storage. We found that with equivalent or even smaller model sizes, our algorithm demonstrates superior performance compared to prior works, illustrating the effectiveness of K-Means based quantization. Lastly, our approach is simple and efficient. Previous studies such as OmniQuant~\cite{shao2023omniquant} and LLM-QAT~\cite{liu2023llmqat} trained auxiliary network components to ensure quantization efficacy. In comparison, our method avoids the necessity of additional training.


\begin{figure}[t!]
  \centering
  \includegraphics[width=0.9\linewidth]{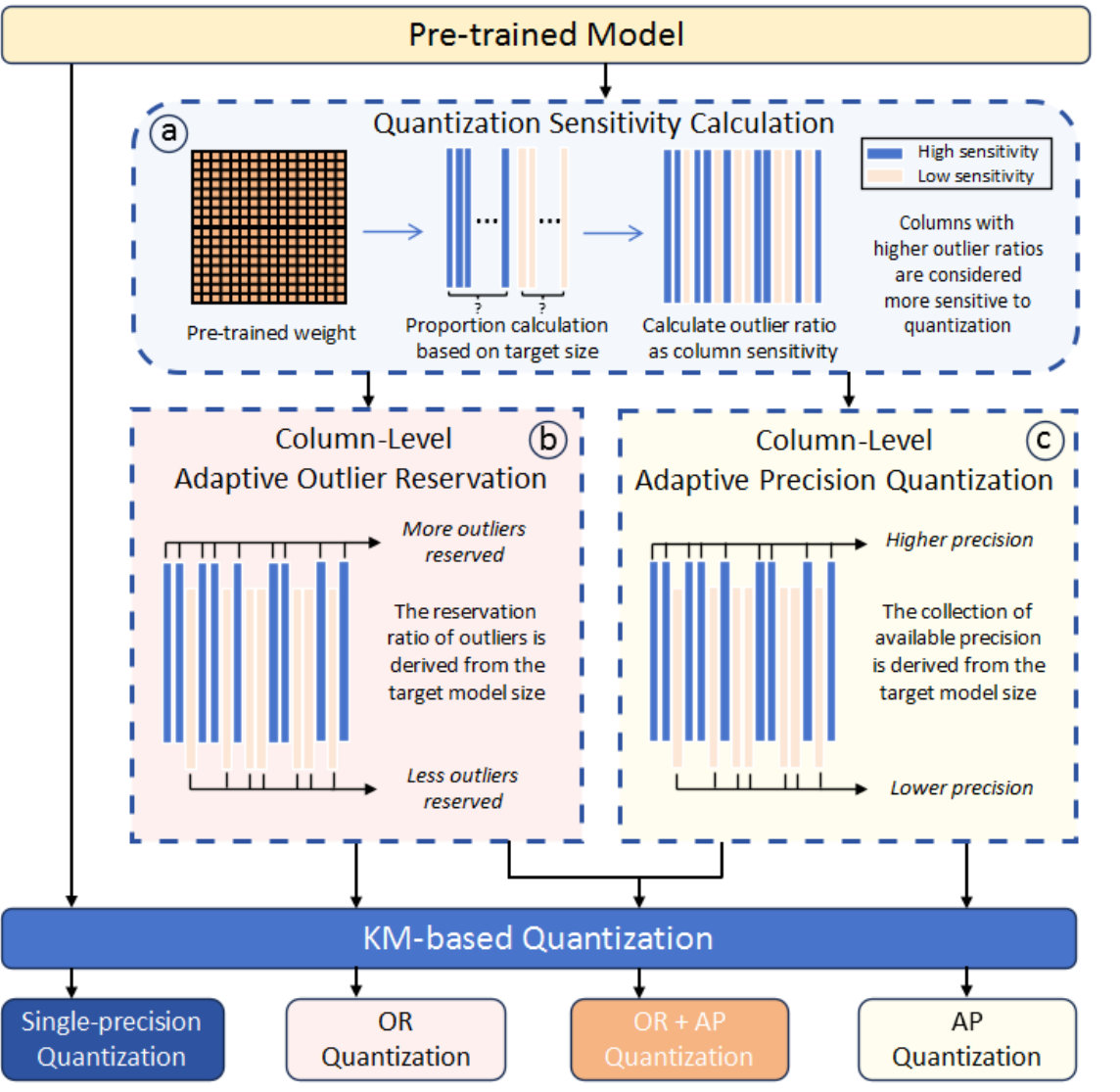}
  \caption{The overall structure of CLAQ: quantized models are obtained from different quantization approaches. The single-precision K-Means based quantization runs without sensitivity calculation. We leverage outlier ratio based quantization sensitivity metric (a) to provide the guidance of the column-level adaptive outlier reservation (OR, b) and column-level adaptive precision (AP, c). The AP and OR strategies are orthogonal to each other.}
  \label{fig:2}
\end{figure}

\subsection{Outlier Ratio Based Quantization Sensitivity Metric}

Previous studies \cite{xiao2023smoothquant, lin2023awq} revealed that outliers in LLMs play a pivotal role in determining the performance of quantized models. Inspired by \cite{yin2023owl}, we hypothesize that the number of outliers within the columns of the weight matrix can act as an indicator of the importance (or sensitivity) of each column  for quantization. Therefore, we adopt outlier ratio as the key guiding metric for LLM quantization.

In this work, we introduce a novel, computationally efficient, and inference-agnostic adaptive precision guidance metric, termed Outlier Order. It can be utilized to allocate adaptive precision (AP) or apply Outlier Reservation (OR) according to the proportion of outliers within each column of the model weight. Columns with a high proportion of outliers, which are considered more important and sensitive to quantization, are allocated higher precision or more outlier reservation budget to preserve model capability, while columns with a lower outlier ratio are assigned lower precision or less reservation budget.

Specifically, for an $i \times j$ weight matrix $W$, the ratio of outliers $R_j$ of the $j$-th column $W_j$ is defined as:

\begin{equation}
R_j = \frac{Card(|W_{j}|>|\overline{W}|\times S)}{i},
\end{equation}

where $Card$ is the counting function, and the numerator represents the number of elements in $W_{j}$ which are greater than the $S$ times of the mean of $W$. $S$ is a scale coefficient for outlier selection, (\textit{e.g.} typically 3, 5, 7, 11, \textit{etc.}). The outlier rate $R_j$ is thus the proportion of parameters which are greater than multiple times of the mean value in the entire matrix $W$. In other words, the larger scale $S$, the greater absolute value required to become an outlier, and the fewer number of outliers picked. Ablation study about the choice of $S$ is conducted in Appendix B. With the help of $R_j$, one can rank matrix columns according to their quantization sensitivities, which facilitates the adaptive selection strategies described below. We name the ranking of $R_j$ as Outlier Order. 

\subsection{Column-Level Adaptive Precision Quantization}

Mixed Precision quantization (MP) involves the integration of multiple quantization levels with varying precisions to mitigate the performance degradation in highly compressed models.  Mixed-precision has been applied to LLMs in previous studies \cite{daillm, li2023lijinhao}. \cite{daillm} leveraged gradient magnitudes of parameters as a guiding principle for determining quantization precision levels. \cite{li2023lijinhao} adopted a conventional criterion based on the relative magnitude of parameters concerning the input. Nonetheless, their metrics of precision allocation can not accurately reflect the importance of parameters and the performances of mixed-precision quantization were unsatisfactory.

To further improve the performance of low-bit quantization, we present a Column-Level Adaptive Precision (AP) strategy directed by Outlier Order, which avoids the heavy computational overhead brought by metric calculation. Additionally, experimental results reveal the AP approach guided by Outlier Order outperforms previous methods with other mixed-precision metrics. To perform the adaptive precision quantization, firstly, the parameters of the matrix are sorted per column based on their outlier ratio as described in section 3.2. Secondly, the bit allocation scheme is determined according to the target compression ratio. For example, a 2.2-bit quantized model is derived by allocating top 10\% outlier-concentrated columns to 4-bit, and allocating 2-bit to the rest. Through AP quantization, the columns with higher sensitivity to quantization are allocated higher precision. Let $B$ be the candidate set of bit-widths; for simplicity, we only describe applying two precision levels in the model (\textit{e.g.}, $B=\{2,4\}$ or $B=\{3,4\}$). The quantization precision for the $j$-th column $P_j$ of the model is denoted as follows:


\begin{equation}
B= \left \{ p_1,p_2 \right \},\quad p_1>p_2, \quad
  \begin{cases}
    P_j = p_1, &\text{if $R_j>T_{AP}$},\\
    P_j = p_2, &\text{Otherwise},
  \end{cases}
\end{equation}

where the threshold $T_{AP}$ is calculated based on the target model size. $T_{AP}$ is a key criterion to ensure that the size of the generated model is close to the desired size limit. For instance of 2.2-bit adaptive quantization, the threshold $T_{AP}$ should be calibrated to the value at the top 10\% of column-wise outlier ratios. With this approach, AP quantization models can be systematically derived to fit any particular size requirement in the range of $B$. Detailed analysis on various bit-width candidates in adaptive precision schemes can be found in Appendix D. Experiments on assigning different bits to different parameter matrices are described in Appendix G.

\subsection{Column-Level Adaptive Outlier Reservation }
As \cite{lin2023awq,lee2024owq, kim2023squeezellm,ashkboos2023quik} have revealed, the preservation of an appropriate fraction of outliers can significantly improve the performance of quantized models. However, previous approaches have simply retained outliers based on the entire parameter matrix and resulted in sub-optimal performances. 


In this work, we propose a Column-Level Adaptive Outlier Reservation (OR) strategy to adaptively preserve full-precision outliers across different columns. We leverage the outlier ratio defined in Sec 3.2 to guide their retention strategy. The analysis of outlier distribution in weight (see Appendix A) illustrates that 90\% of outliers are concentrated within the top 10\% of columns, leading us to adopt a retention policy that preserves a greater number of outliers in the initial 10\% of columns, while conserving fewer outliers in the remaining 90\%. 

To apply the OR strategy, firstly, we pick top 10\% quantization-sensitive columns with Outlier Order metric. Then we preserve a higher proportion $o_1$ of outliers unquantized in top 10\% columns with higher outlier ratios, and maintain a lower outlier retention ratio $o_2$ in the rest. In each column, the same number of the largest and smallest parameters are reserved as outliers. The total number of outliers reserved is calculated by the target size of compression. The choice of $o_1$ and $o_2$ decides the proportion of outliers retained across columns. Grid search on $o_1$ and $o_2$ are discussed in Appendix C. Column-Level Adaptive Outlier Reservation (OR) ratio $O_j$ for the $j$-th column is presented below:

\begin{equation}
OR= \left \{ o_1,o_2 \right \}, \quad o_1>o_2, \quad
  \begin{cases}
    O_j = o_1, &\text{if $R_j>T_{OR}$},\\
    O_j = o_2, &\text{Otherwise},
  \end{cases}
\end{equation}

where $OR$ denotes the set of retained outlier ratios. $T_{OR}$ represents the threshold of choosing the top 10\% columns of higher outlier ratios. Columns with higher outlier ratio will be assigned more FP16 outlier reservation. Same proportion of outliers is reserved in all matrices to control the size of the compressed model. The advantages of this approach are threefold: First, by retaining more full-precision parameters in quantization-susceptible columns, model performance is better preserved in quantization. Second, the metrics employed for outlier reservation align with those used for generating adaptive precision quantization, avoiding redundant computational overhead. Thirdly, the implementation of outlier reservation and adaptive precision quantization are capable of being employed concurrently. Our fusion model (AP+OR) outperforms other PTQ methods with similar memory footprints.

\section{Experiments and Results}

\subsection{Experimental Settings}

\paragraph{Models}
The experiments are conducted on popular public LLMs. We adopted LLaMA \cite{touvron2023LLaMA} and Yi \cite{young2024yi} as the baseline models, known for their superior performances in English and Chinese language tasks, respectively. The pre-trained base versions are employed in our experiments. 

\paragraph{Datasets}
We employ widely-used datasets C4 \cite{raffel2020c4} and WiKiText2 \cite{merity2016wiki2} as perplexity test sets. To evaluate downstream task capabilities, we tested on a diverse range of zero-shot benchmarks, including WinoGrande \cite{sakaguchi2021winogrande} , PiQA \cite{bisk2020piqa} , HellaSwag \cite{zellers2019hellaswag}, ARC (comprising ARC-easy and ARC-challenge) \cite{clark2018think} and BoolQ \cite{clark2019boolq}. Additionally, C4 was chosen as the calibration dataset for tuning our quantization process. It is noteworthy to emphasize that the choice of calibration data might have an impact on experimental outcomes, as detailed in Appendix H.

\begin{table}[t]
\centering
\caption{Results of perplexity on WikiText2 \cite{merity2016wiki2} and C4 \cite{raffel2020c4} of our proposed CLAQ and other methods. CLAQ models with * are fusion models employing column-level adaptive precision and outlier reservation techniques. The detail of fusion model setups is described in Appendix F. $g=128$ means group size of quantization is 128, followed by the equivalent bit-width in italics. The results of LLaMA-2 series and Yi-34B models can be found in Appendix E.}
\label{table:1}
\resizebox{\textwidth}{!}{
    \begin{tabular}{@{}cccccccccc@{}}
    \toprule
\textbf{Method}      & \textbf{\# Bits}      & \textbf{\begin{tabular}[c]{@{}c@{}}LLaMA1-7B\\ WikiText2↓/C4↓\end{tabular}} & \textbf{\begin{tabular}[c]{@{}c@{}}LLaMA1-13B\\ WikiText2↓/C4↓\end{tabular}} & \textbf{\begin{tabular}[c]{@{}c@{}}LLaMA1-30B\\ WikiText2↓/C4↓\end{tabular}} & \textbf{\begin{tabular}[c]{@{}c@{}}LLaMA1-65B\\ WikiText2↓/C4↓\end{tabular}} \\
\midrule
 --     & 16 & 5.63/7.08 & 5.02/6.61 & 4.04/5.97 & 3.49/5.61   \\
RTN\cite{frantar2022gptq}     & 4  & 6.43/7.93 & 5.55/6.98 & 4.57/6.34 & 3.87/5.85   \\
GPTQ\cite{frantar2022gptq}    & 4  & 5.99/7.43 & 5.26/6.83 & 4.35/6.20 & 3.77/5.79   \\
LLM-QAT\cite{liu2023llmqat} & 4  & 10.9/7.4  & 9.6/\textbf{6.5}   & 7.3/6.5   & --          \\
AWQ\cite{lin2023awq}     & 4  & 6.08/7.52 & 5.34/6.86 & 4.39/6.17 & 3.76/5.77   \\
OmniQuant\cite{shao2023omniquant}  & 4 & 5.86/7.34 & 5.21/6.76 & 4.25/6.11 & 3.71/5.73  \\
SqueezeLLM\cite{kim2023squeezellm}  & 4.05 & 5.79/7.21 & 5.18/6.71 & 4.22/6.06 & 3.76/5.69   \\
SpQR\cite{dettmers2023spqr}  & 3.94/3.96/3.89/3.90  & 5.87/7.28  & 5.22/6.72 & 4.25/6.08         & 3.68/5.70 \\
\textbf{CLAQ}  & 4 & \textbf{5.78}/\textbf{7.21} & \textbf{5.15}/6.70 & \textbf{4.17}/\textbf{6.06} & \textbf{3.62}/\textbf{5.69} \\
\midrule
GPTQ\cite{frantar2022gptq}   & 3 & 8.0/9.54 & 6.54/8.15 & 5.59/7.29 & 4.94/6.69   \\
AWQ\cite{lin2023awq}    & 3 & 11.88/13.26 & 7.45/9.13 & 10.07/12.67 & 5.21/7.11 \\
OmniQuant\cite{shao2023omniquant}   & 3 & 6.49/8.19 & 5.68/7.32 & \textbf{4.74}/6.57                         & \textbf{4.04}/6.07 \\
\textbf{CLAQ}  & 3 & \textbf{6.47/7.87} & \textbf{5.61/7.14}                & 4.79/\textbf{6.49} & 4.11/\textbf{6.00} \\
\midrule
AWQ\cite{lin2023awq}  & 3 ($g=128$,\textit{3.15}) & 6.46/7.92 & 5.51/7.07 & 4.63/6.37 & 3.99/5.94  \\
OmniQuant\cite{shao2023omniquant}  & 3 ($g=128$,\textit{3.15}) & 6.15/7.75 & 5.44/7.05 & 4.56/6.37 & 3.94/5.93                                                                    \\
SqueezeLLM\cite{shao2023omniquant}  & 3.24 & 6.13/7.56 & 5.45/6.92 & 4.44/6.23 & 3.88/5.84   \\
\textbf{CLAQ*}  & 3.12 & \textbf{5.97/7.40} & \textbf{5.27/6.83}  & \textbf{4.35/6.18}  & \textbf{3.75/5.78}   \\
\textbf{CLAQ*}  & 3.23 & \textbf{5.95/7.37} & \textbf{5.24/6.81}             & \textbf{4.33/6.16} & \textbf{3.74/5.76} \\
\midrule
GPTQ\cite{frantar2022gptq} & 2 & 2148.66/691.25 & 8730.02/1467.84 & 508.02/188.73                 & \textbf{58.34/39.57} \\

\textbf{CLAQ} & 2 & \textbf{27.64/24.37} & \textbf{21.05/19.69}        & \textbf{15.37/13.43} & 97.16/48.95 \\
\midrule
AWQ\cite{lin2023awq}  & 2 ($g=64$,\textit{2.28}) & 2.5e5/2.8e5 & 2.7e5/2.2e5 & 2.3e5/2.3e5 
& 7.4e4/7.4e4  \\
OmniQuant\cite{shao2023omniquant}  & 2 ($g=64$,\textit{2.28})  & 8.90/11.78 & 7.34/9.75 & 6.59/8.65 & 5.65/7.60 \\
decoupleQ\cite{guo2024decoupleq}  & 2 ($g=64$,\textit{2.28})  & 8.18/- & 6.96/- & 5.81/- & 5.07/- \\
\textbf{CLAQ*} & 2.12 & \textbf{7.57/8.87} & \textbf{6.41/7.92}          & \textbf{5.40/7.05}  & \textbf{4.70/6.46} \\
\textbf{CLAQ*}  & 2.24 & \textbf{6.93/8.35} & \textbf{5.99/7.52}          & \textbf{5.03/6.74} & \textbf{4.41/6.22}    \\                      

\bottomrule
\end{tabular}}
\end{table}

\renewcommand{\floatpagefraction}{.9}
\begin{table}[ht]

\centering
\caption{The evaluation result of Zero-Shot accuracy of our proposed CLAQ and other methods. CLAQ models with * are fusion models employing column-level adaptive precision and outlier reservation techniques. The results of LLaMA-2 series, Yi-34B model and experiments on other bit-width can be found in Appendix E.}
\resizebox{\textwidth}{!}{
    \begin{tabular}{@{}ccccccccccc@{}}
    \toprule
\textbf{Model} &\textbf{Method} & \textbf{\# Bits}  & \textbf{PIQA} & \textbf{Arc-e} &\textbf{Arc-c} & \textbf{BoolQ} & \textbf{HellaSwag}& \textbf{Winogrande} &\textbf{Avg↑} \\
\midrule
 & FP16 & 16 & 79.11 & 75.25 & 44.62 & 75.11 & 76.20 & 70.01 & 70.38 \\
 & LLM-QAT\cite{liu2023llmqat} & 4 & 78.3 & 70.0 & 45.0 & 75.5 & 74.0 & 69.0 & 68.63 \\
 & GPTQ\cite{frantar2022gptq} & 4 & 78.67 & 73.23 & 42.83 & 74.16 & 75.07 & 70.48 & 69.07 \\
 & OmniQuant\cite{shao2023omniquant} & W6A6 & 77.09 & 51.89 & 40.87 & 72.53 & 71.61 & 65.03 & 63.17 \\
   & SpQR\cite{dettmers2023spqr} & 4.63 & 78.45 & 67.13 & 38.23 & - & 56.01 & 67.48 & 61.46 \\
\multirow{-3}{*}{\begin{tabular}[c]{@{}c@{}}LLaMA1-7B\end{tabular}} & \textbf{CLAQ} & 4 & 78.29 & 75.29 & 43.25 & 74.70 & 75.26 & 70.32 & \textbf{69.52} \\
\cmidrule(l){2-10}
 & GPTQ\cite{frantar2022gptq} & 2 & 52.50 & 25.76 & 26.96 & 41.87 & 25.77 & 52.49 & 37.56 \\
& \textbf{CLAQ*} & 2.12 & 75.73 & 68.56 & 36.43 & 70.73 & 67.05 & 67.25 & \textbf{64.29} \\
\midrule
 & FP16 & 16 & 80.14 & 77.40 & 47.70 & 77.92 & 79.09 & 72.85 & 72.52 \\
 & LLM-QAT\cite{liu2023llmqat} & 4 & 79.4 & 72.8 & 52.0 & 77.7 & 77.7 & 71.5 & 71.85 \\
 & GPTQ\cite{frantar2022gptq} & 4 & 79.60 & 76.60 & 47.87 & 76.45 & 77.89 & 71.90 & 71.72 \\
 & OmniQuant\cite{shao2023omniquant} & W6A6 & 78.40 & 57.28 & 42.91 & 67.00 & 75.82 & 68.27 & 64.95 \\
   & SpQR\cite{dettmers2023spqr} & 4.63 & 78.94 & 74.37 & 43.17 & - & 59.02 & 69.77 & 65.05 \\
\multirow{-3}{*}{\begin{tabular}[c]{@{}c@{}}LLaMA1-13B\end{tabular}} & \textbf{CLAQ} & 4 & 79.65 & 76.94 & 48.04 & 77.74 & 78.58 & 72.85 & \textbf{72.30} \\
\cmidrule(l){2-10}
& GPTQ\cite{frantar2022gptq} & 2 & 52.18 & 25.88 & 28.41 & 41.35 & 25.67 & 49.64 & 37.19 \\
& \textbf{CLAQ*} & 2.12 & 77.80 & 72.22 & 41.21 & 69.42 & 71.86 & 69.14 & \textbf{66.94} \\
\midrule
 & FP16 & 16 & 82.26 & 80.43 & 52.90 & 82.69 & 82.63 & 75.85 & 76.13 \\
 & LLM-QAT\cite{liu2023llmqat} & 4 & 81.0 & 79.4 & 56.8 & 81.8 & 81.8 & 75.1 & \textbf{75.98} \\
 & GPTQ\cite{frantar2022gptq} & 4 & 81.12 & 76.81 & 50.00 & 79.36 & 81.43 & 74.66 & 73.90 \\
 & OmniQuant\cite{shao2023omniquant} & W6A6 & 79.81 & 58.79 & 45.22 & 68.38 & 78.95 & 72.21 & 67.23 \\
   & SpQR\cite{dettmers2023spqr} & 4.69 & 81.01 & 76.05 & 47.18 & - & 62.50 & 72.93 & 67.93 \\
\multirow{-3}{*}{\begin{tabular}[c]{@{}c@{}}LLaMA1-30B\end{tabular}} & \textbf{CLAQ} & 4 & 81.18 & 80.39 & 54.10 & 82.17 & 81.91 & 75.85 & 75.93 \\
\cmidrule(l){2-10}
 & GPTQ\cite{frantar2022gptq} & 2 & 53.16 & 27.74 & 26.79 & 43.46 & 27.10 & 48.46 & 37.79 \\
 &\textbf{ CLAQ*} & 2.12 & 80.03 & 76.73 & 47.10 & 79.51 & 76.55 & 73.64 & \textbf{72.26} \\
\midrule
 & FP16 & 16 & 82.26 & 81.36 & 55.55 & 84.83 & 84.12 & 77.27 & 77.57 \\
 & GPTQ\cite{frantar2022gptq} & 4 & 82.48 & 80.89 & 54.01 & 83.55 & 83.47 & 76.68 & 76.85 \\
 & OmniQuant\cite{shao2023omniquant} & W6A6 & 81.01 & 58.12 & 46.33 & 80.64 & 79.91 & 75.69 & 70.28 \\
  & SpQR\cite{dettmers2023spqr} & 4.71 & 81.56 & 75.25 & 46.93 & - & 63.76 & 76.95 & 68.89 \\
\multirow{-2}{*}{\begin{tabular}[c]{@{}c@{}}LLaMA1-65B\end{tabular}} & \textbf{CLAQ} & 4 & 82.43 & 81.52 & 54.69 & 83.85 & 83.56 & 76.40 & \textbf{77.08} \\
\cmidrule(l){2-10}
 & GPTQ\cite{frantar2022gptq} & 2 & 53.59 & 28.62 & 27.22 & 54.80 & 30.09 & 51.62 & 40.99 \\
 & \textbf{CLAQ*} & 2.12 & 80.20 & 78.96 & 50.17 & 79.14 & 79.39 & 75.30 & \textbf{73.86} \\
\bottomrule
\label{table:2}
\end{tabular}}
\end{table}

\paragraph{Implementation Details}

To implement CLAQ, we leveraged the accelerated K-Means clustering algorithm provided by the scikit-learn-intelex library \footnote{https://github.com/intel/scikit-learn-intelex (Apache-2.0 license)}. Our experiments were built upon the GPTQ framework \footnote{https://github.com/IST-DASLab/gptq (Apache-2.0 license)}. For benchmarking purposes, the LM-Evaluation-Harness toolkit \footnote{https://github.com/EleutherAI/lm-evaluation-harness (MIT License)} served as the assessment platform. All experiments were conducted on a server with 8 NVIDIA A100 GPUs, each with 80GB capacity. Notably, a single NVIDIA A100 GPU is sufficient for the majority of our experiments except evaluating the LLaMA-65B model on zero-shot tasks, where two GPUs were used to accommodate its memory demands. Our contribution was mainly confined to the quantization of model weights, with activations left unquantized. Unless otherwise specified within the experimental results table, a 16-bit precision was maintained for activations. Comprehensive details regarding the hyper-parameters employed in our experiments can be found in Appendix F.

\subsection{Main Results}

The main results of our experiments are presented in Table \ref{table:1} and Table \ref{table:2}. Table \ref{table:1} shows the evaluations on perplexity, while the performance of zero-shot tasks is given in Table \ref{table:2}. It can be clearly observed that CLAQ consistently outperforms existing methods in various scenarios. Notably, in case of 3-bit and 4-bit single-precision quantization with a comparable codebook size configuration, our method generally outperforms the baselines examined.

For lower bit-width, we employ the fusion approach of AP and OR. Results show that with an additional 0.1-bit memory footprint, model accuracy is significantly boosted. Especially for compression around 2-bit (2.1 bits in our case), our method exhibits a consistent and substantial lead over all existing methods, achieving the state-of-the-art (SOTA) performance (significant difference at 0.01 confidence level).

Also, the zero-shot evaluation results demonstrate the superiority of our method. CLAQ outperforms other baselines in most cases. Specifically, in the case of 4-bit quantization, our solution closely matches the performance of full-precision models. For low-bit scenarios, our method significantly improve the accuracy of the quantized model in zero-shot tasks, and pushing the limits of large language model quantization under low-bit conditions. More zero-shot task results on LLaMA-2 series and Yi-34B models can be found in Appendix E.


\subsection{Ablation Study}
\subsubsection{Adaptive Precision Quantization}
Table \ref{table:3} presents a comparison of different Adaptive Precision  (AP) quantization methods across various precision levels. Inspired by \cite{frantar2023sparsegpt}, we developed a magnitude-to-parameter method to guide the mixed-precision allocation. Experimental results demonstrate that our proposed column-level adaptive precision quantization strategy significantly outperforms other mixed-precision methods under comparable size. The improvements can be attributed to the superiority of our Outlier Order metric. According to the outlier distribution analysis (see Appendix A), a minority of columns exhibit higher sensitivity to quantization. Consequently, allocating additional precision on preserving these critical columns is an effective way to improve the performance of quantized models.  

\subsubsection{Outlier Reservation}
Table \ref{table:4} lists the results of the fixed and Outlier Reservation (OR) strategy, highlighting its superiority over static outlier preservation approaches. Remarkably, with an equal amount of increase in model sizes, the performance gain from retaining more outliers exceeds that from adaptive precision strategy. This result suggests the importance of outlier retention, indicating that directly preserving outliers at full precision is superior to quantizing them with a higher precision. 

\begin{table}[ht]
\centering
\caption{Ablation study results for our proposed column-level adaptive precision quantization. \dag \enspace denotes a column-wise mixed-precision quantization with the metric of activation-to-weight ratio proposed in \cite{frantar2023sparsegpt}.}
\label{table:3}
\resizebox{\linewidth}{!}{
\begin{tabular}{lcccccccccc}
\toprule
\multicolumn{1}{c}{\textbf{Model}} & \textbf{\# Bits} & \textbf{WikiText2↓} & \textbf{C4↓}  & \textbf{PIQA} & \textbf{Arc-e} &\textbf{Arc-c} & \textbf{BoolQ} & \textbf{HellaSwag}& \textbf{Winogrande} &\textbf{Avg↑} \\
\midrule
LLaMA1-7B       & 16               & 5.63     & 7.08   & 79.11   & 75.25  & 44.62   & 75.11        & 76.20  & 70.01  & 70.38   \\
CLAQ  & 3   & 6.47     & 7.87  & 78.23    & 72.05   & 41.12  & 72.93  & 72.05            & 67.71   & 67.35      \\
CLAQ     & 2     & 27.64     &24.37 & 62.89    & 31.43   & 25.51           & 61.52       & 28.69   & 51.85   & 43.65   \\

\midrule
 +MP\dag   & 2.5   & 8.63  & 9.98    & 75.35         & 65.15   & 36.51   & 69.08   & 62.96   & 64.71     & \textbf{62.29}   \\
 \textbf{+AP(ours)}   & 2.5    & \textbf{8.43}       & \textbf{9.78}   & 75.63  & 68.10 & 35.24  & 66.39   & 63.82   & 64.40        & 62.26    \\
\midrule
 +MP\dag  & 2.2   & 11.60             & 12.70    & 71.00  & 57.91  & 29.43   & 65.22   & 51.07   & 59.03     & 55.61  \\
 \textbf{+AP(ours)}   & 2.2    & \textbf{10.14}  & \textbf{11.46} & 73.39   & 64.10   & 31.91   & 63.52   & 60.08  & 62.83  & \textbf{59.31}   \\
\midrule
 +MP\dag  & 2.1   & 16.42    & 16.57    & 66.86      & 44.02    & 26.53    & 62.56    & 44.03     & 53.59     & 49.60 \\
 \textbf{+AP(ours)} & 2.1   & \textbf{11.06}   & \textbf{12.34} & 72.42  & 63.17    & 31.06  & 64.92  & 58.33    & 61.25  & \textbf{58.53}        \\
\bottomrule
\end{tabular}}
\end{table}

\begin{table}[ht]
\centering
\caption{Ablation study results for our proposed column-level adaptive outlier reservation. Outlier fix refers to keeping a fixed proportion of outliers in each column.}
\label{table:4}

\resizebox{\linewidth}{!}{
\begin{tabular}{lccccccccccc}
\toprule
\multicolumn{1}{c}{\textbf{Model}} & \textbf{\# Bits} &  \textbf{WikiText2↓} & \textbf{C4↓}  & \textbf{PIQA} & \textbf{Arc-e} &\textbf{Arc-c} & \textbf{BoolQ} & \textbf{HellaSwag}& \textbf{Winogrande} &\textbf{Avg↑} \\
\midrule
\multicolumn{1}{c}{LLaMA1-7B}       & 16                & 5.63  & 7.08    & 79.11  & 75.25     & 44.62   & 75.11         & 76.20    & 70.01  & 70.38     \\
CLAQ   & 2    & 27.64   & 24.37 & 62.89   & 31.43   & 25.51   & 61.52 & 28.69            & 51.85    & 43.65   \\
\midrule
+Outlier fix   & 2.28         & 7.49   & 8.87      & 74.70   & 68.56  & 37.88   & 71.52  & 67.03  & 66.69   & 64.40    \\
\textbf{+OR(ours)}  & 2.28      & \textbf{6.88}    & \textbf{8.35}  & 77.75     & 69.36   & 39.76    & 74.40   & 70.62   & 67.09   & \textbf{66.50}    \\
\midrule
+Outlier fix  & 2.14   & 8.31   & 9.61    & 75.19         & 66.41   & 35.32  & 71.40    & 63.12    & 66.14    & 62.93   \\
\textbf{+OR(ours)}  & 2.14   & \textbf{7.22}   & \textbf{8.64}  & 76.28     & 69.02      & 37.88    & 73.70    & 68.14    & 68.11    & \textbf{65.52}  \\
\bottomrule
\end{tabular}}
\end{table}

\section{Limitations and Future Works}
The main limitations of our work lie in the simplification of outlier retention and adaptive precision search strategies. We adopt two-level precision in AP for the convenience of CUDA kernel development. However, when the assignable precision bit-width exceeds the equivalent of 0.5 bit (i.e., >2.5bit, >3.5bit), the adaptive precision strategy outlined in our method may not represent the optimal approach. The potential for generating superior adaptive precision configurations is discussed in Appendix G, where we propose a heuristic adaptive precision search algorithm. Our future work is twofold: (1) We will continue to investigate the accomplishment of superior adaptive precision quantization configurations through efficient search algorithms. (2) We are developing customized CUDA kernels to support efficient computation of the CLAQ approach on GPUs. Compared with the AP quantization, dynamically preserving outliers per column is more challenging in deployment. We are actively refining the underlying algorithms, and the code will be released soon in the future.

\section{Conclusion}
In this paper, we have introduced a novel Column-Level Adaptive weight Quantization (CLAQ) framework, which integrates column-level adaptive precision quantization and outlier reservation techniques. Guided by the column-wise outlier proportion metric, the CLAQ method achieves efficient compression of LLMs while maximizing the preservation of model accuracy. Extensive evaluations on public LLM models confirm the superiority of CLAQ in enhancing performance across varied bit-widths, notably in ultra-low bit cases. The results lend help to making LLMs more efficient and accessible in resource-constrained applications.

\newpage



\medskip

\small


\newpage
\appendix

\section{Analysis on outlier ratio in LLM}
\begin{figure}[H]
  \centering
  \includegraphics[width=0.54\linewidth]{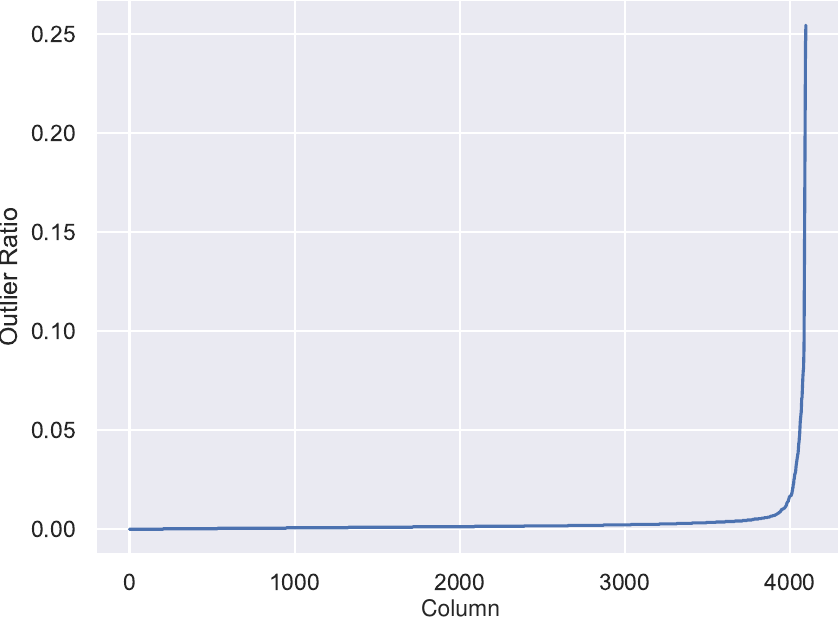}
  \caption{The sorted outliers ratio in a self-attention matrix of LLaMA1-7B, most columns contain few outliers.}
  \label{fig:3}
\end{figure}

In Section 3.2, we underscored the significance of outliers in the context of Large Language Model (LLM) compression. Here, we present concrete examples to demonstrate the efficacy of our approach. Initially, we observe notable variations in outlier prevalence among different columns within each parameter matrix. Figure \ref{fig:3} illustrates the statistics of outliers detected when applying an outlier standard deviation $S=7$ to the layer \texttt{layers.0.self\_attn.o\_proj.weight}. The findings highlight that in the LLaMA architecture, outliers are confined to a minority of columns, with a striking imbalance in their distribution across columns. Consequently, rationing the limited precision to those columns exhibiting a higher concentration of outliers is advocated as a strategy to maximize performance gains.

\begin{figure}[H]
  \centering
  \includegraphics[width=0.45\linewidth]{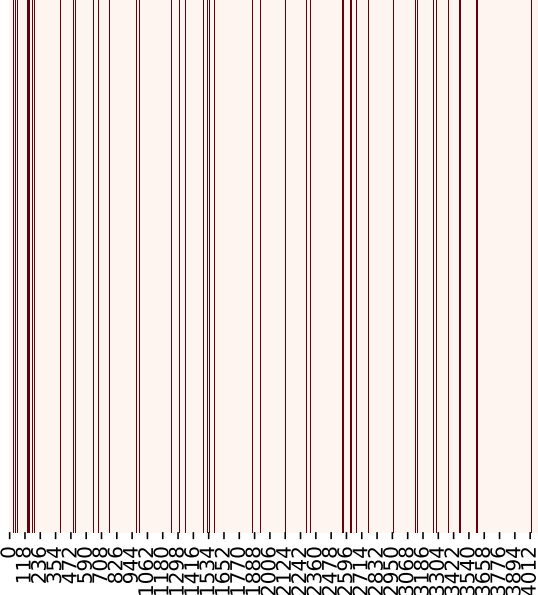}
  \caption{The position of columns with higher outlier ratio in matrix. Columns with dark colour are the top 10\% outlier concentrated columns.}
  \label{fig:4}
\end{figure}

Figure \ref{fig:4} displays the top 10\% columns with higher outlier ratio in the same matrix, which are more evenly distributed with no apparent pattern. This intuitive evidence highlights the heterogeneity between these columns, further emphasizing the effectiveness for selective precision assignments.

\begin{figure}[H]
  \centering
  \includegraphics[width=0.9\linewidth]{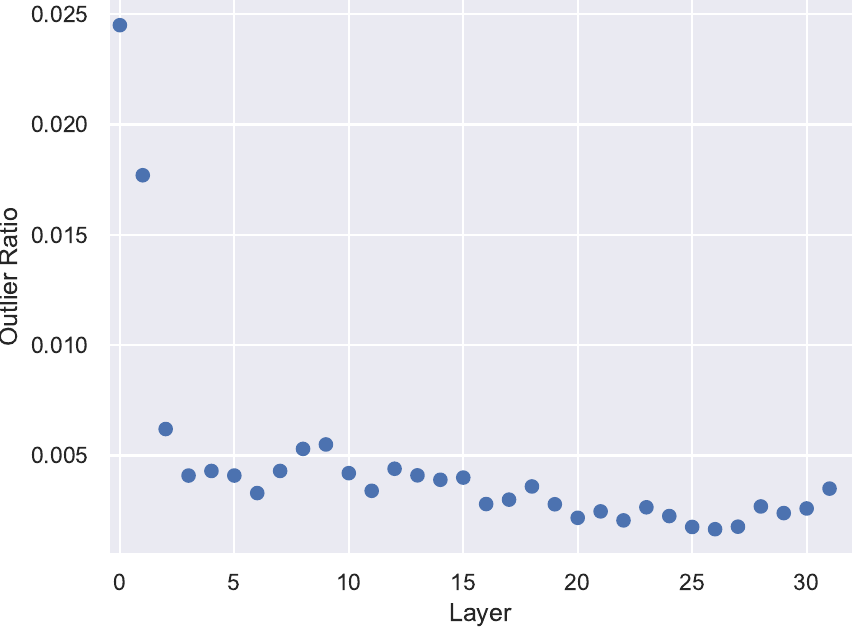}
  \caption{The overall outlier ratio of 32 layers in LLaMA1-7B.}
  \label{fig:5}
\end{figure}

Extending the analysis to the entire model, Figure \ref{fig:5} summarizes the quantization outcomes for all 32 decoder layers in LLaMA1-7B. From a holistic model perspective, the initial layers exhibit a disproportionately high outlier incidences. Aligning with our hypothesis, this suggests that these layers hold increased significance. Considering the disparities between layers, we have explored a heuristic-based search strategy for identifying potentially improved adaptive precision allocation schemes, details of this approach are elaborated in Appendix G.

\section{Ablation study on outlier definition}
The selection of the criterion for defining outliers is integral to the efficacy of the Outlier Order search process. To this end, we conducted an ablation study examining the impact of varying the outlier standard, denoted as $S$. Our findings reveal that $S=13$ emerges as the optimal setting, striking a balance where it neither lowers the threshold excessively to include an abundance of values nor sets the bar so high that certain columns lack any outliers. This configuration ensures accurate identification of meaningful outliers that significantly influence the performance of the model quantization strategy.

\begin{table}[H]
\centering
\caption{Ablation study on the standard of outlier definition on LLaMA1-7B. CLAQ* denotes the column-level adaptive precision quantization is applied.}
\begin{tabular}{ccccc}
\toprule
\textbf{Method} & \textbf{\# Bits} & \multicolumn{1}{c}{\textbf{Outlier Standard}} & \textbf{WikiText2↓} & \textbf{C4↓} \\
\midrule
CLAQ* & 2.2 & $S=1$ & 16.39 & 16.47 \\
CLAQ* & 2.2 & $S=3$ & 13.44 & 14.52 \\
CLAQ* & 2.2 & $S=5$ & 10.76 & 12.07 \\
CLAQ* & 2.2 & $S=7$ & 10.58 & 11.71 \\
CLAQ* & 2.2 & $S=9$ & 10.37 & 11.64 \\
CLAQ* & 2.2 & $S=11$ & 10.25 & 11.58 \\
CLAQ* & 2.2 & $S=13$ & \textbf{10.14} & \textbf{11.46} \\
CLAQ* & 2.2 & $S=15$ & 10.32 & 11.53 \\
CLAQ* & 2.2 & $S=17$ & 10.34 & 11.51 \\
\bottomrule
\end{tabular}
\end{table}

\section{Ablation study on hyper parameter in outlier reservation}

Observations have indicated that the top 10\% of columns harbor approximately 90\% of the outliers, prompting an investigation into the optimal strategy for allocating outlier retention within these columns with top outlier ratio versus the remaining 90\%. We keep this proportion of columns to retain adaptively outliers in our experiments. The challenge lies in determining the most efficacious ratio for preserving outliers in these two segments. Various configurations were trialed in pursuit of the most suitable distribution. 

We did grid search on the choice of outlier reservation ratio. We found 3 promising configuration settings: \textbf{Setting 1: 19\% for high outlier ratio columns, 81\% for 90\% low outlier ratio columns. \textbf{Setting 2:} 28\% for high outlier ratio columns, 72\% for 90\% low outlier ratio columns. \textbf{Setting 3:} 37\% for high outlier ratio columns, 63\% for 90\% low outlier ratio columns.} Experimental results shows setting 3 outperforms in terms of PPL, while setting 2 obtained the higher accuracy on zero-shot tasks than other settings. The performance gap on PPL is more obvious than zero-shot accuracy. Finally, striking the balance between downstream task accuracy and perplexity, \textbf{we decide to choose \textbf{Setting 2} for our main experiments.}

\begin{table}[H]
\centering
\caption{Experimental results of ablation study conducted on different hyper-parameter settings about outlier reservation proportion. }
\begin{tabular}{lccccc}
\toprule
\textbf{Model} & \textbf{\# Bits} & \multicolumn{1}{c}{\textbf{Exp. Config}} & \textbf{WikiText2↓} & \textbf{C4↓} & \textbf{Zero-Shot Avg↑} \\
\midrule
LLaMA1-7B & 16 & \multicolumn{1}{c}{--} & 5.63 & 7.08 & 70.38\\
\midrule
+OR & 2.28 & Setting1 & 7.01 & 8.47 & 65.98\\
+OR & 2.28 & Setting2 & 6.88 & 8.35 & \textbf{66.50}\\
+OR & 2.28 & Setting3 & \textbf{6.77} & \textbf{8.31} & 65.21 \\
\midrule
+OR & 2.14 & Setting1 & 7.55 & 8.93 & 64.31\\
+OR & 2.14 & Setting2 & 7.22 & 8.64 & \textbf{65.52}\\
+OR & 2.14 & Setting3 & \textbf{7.00} & \textbf{8.49} & 65.30\\
\bottomrule
\end{tabular}
\end{table}

\section{Ablation study on adaptive precision}

With the objective of facilitating deployment at the hardware level, our approach adopts a dual-precision quantization scheme. A dilemma arises when targeting a quantization granularity between 2 to 3 bits: given an equivalent model size constraint, is it more beneficial to incorporate 2+3 mixture, or 2+4 mixture? To address this question, we designed a series of experiments under three distinct outlier standard conditions. The outcomes of which are summarized in Table \ref{table:7}. The outcomes consistently lead to a single, definitive conclusion: opting for fewer columns at 4 bits is preferable. This finding aligns with observations detailed in Appendix A, which underscores that columns with a high outlier count are relatively rare. Prioritizing higher precision for these selected columns proves crucial in achieving superior overall quantization efficacy.

\begin{table}[H]
\centering
\caption{Experimental results of ablation study on LLaMA1-7B about adaptive precision candidate bit-width selection.}
\begin{tabular}{cccccc}
\toprule
\textbf{Method} & \textbf{\# Bits } & \textbf{Bits in AP}  & \textbf{Outlier Standard} & \textbf{WikiText2↓} & \textbf{C4↓} \\
\midrule
AP & 2.1 & 2\&3 &$S=5$ & 12.35 & 13.77 \\
AP & 2.1 & 2\&4 &$S=5$ & \textbf{12.21} & \textbf{13.21} \\
\midrule
AP & 2.1 & 2\&3 &$S=9$ & 11.89 & 13.07 \\
AP & 2.1 & 2\&4 &$S=9$ & \textbf{11.52} & \textbf{12.73} \\
\midrule
AP & 2.1 & 2\&3 & $S=13$ & 11.67 & 12.78 \\
AP & 2.1 & 2\&4 & $S=13$ & \textbf{11.06} & \textbf{12.34} \\
\bottomrule
\label{table:7}
\end{tabular}
\end{table}

\section{More experimental results}

More experiments on LLaMA2 and Yi are shown in the following tables. The results are consistent with the conclusions in the main results section. Our proposed CLAQ outperformed other quantization method in most settings; especially, CLAQ with AP and OR achieves SOTA performance under the low-bit condition.

\begin{table}[H]
\centering
\caption{Experimental results of perplexity in LLaMA-2-Base models. CLAQ models with * are fusion models employing column-level adaptive precision and outlier reservation
techniques. $g=128$ means group size of quantization is 128, followed by the equivalent bit-width in italics.}
\label{table:8}

\begin{tabular}{cccc}
\toprule
\textbf{Method}      & \textbf{\# Bits}      & \textbf{\begin{tabular}[c]{@{}c@{}}LLaMA2-7B\\ WikiText2↓/C4↓\end{tabular}} & \textbf{\begin{tabular}[c]{@{}c@{}}LLaMA2-13B\\ WikiText2↓/C4↓\end{tabular}} \\
\midrule
-   & 16 & 5.46/6.97 & 4.88/6.46  \\
GPTQ & 4 & 5.82/7.36 & 5.13/6.70   \\
AWQ  & 4 & 6.15/7.68 & 5.12/6.74   \\
AWQ  & 4 ($g=128$,\textit{4.15}) & 5.62/7.13 & 4.97/\textbf{6.56}  \\
OmniQuant & 4 & 5.74/7.35 & 5.02/6.65  \\
OmniQuant & 4 ($g=128$,\textit{4.15}) & \textbf{5.58}/\textbf{7.12} & \textbf{4.95}/\textbf{6.56}  \\
\textbf{CLAQ}  & 4 & 5.62/7.13 & 5.00/\textbf{6.56}    \\
\midrule
GPTQ & 3 & 8.49/9.92 & 6.40/8.00   \\
AWQ  & 3 & 24.00/23.85 & 10.45/13.07  \\
OmniQuant & 3 & 6.58/8.65 & 5.58/7.44  \\
\textbf{CLAQ} & 3 & \textbf{6.39/7.80} & \textbf{5.46/7.03}   \\
\midrule
AWQ & 3 ($g=128$,\textit{3.15}) & 6.24/7.84 & 5.32/6.94  \\
OmniQuant   & 3 ($g=128$,\textit{3.15}) & 6.03/7.75 & 5.28/6.98   \\
\textbf{CLAQ*}  & 3.12 & \textbf{5.81/7.34}   & \textbf{5.15/6.72}      \\
\textbf{CLAQ*}  & 3.23 & \textbf{5.79/7.31} & \textbf{5.12/6.70}   \\
\midrule
GPTQ & 2 & Nan/2477.60 & \textbf{1832.06/349.17}  \\
\textbf{CLAQ} & 2 &\textbf{ 7579.07/1412.06} & 4853.96/3404.98                                                     \\
\midrule
AWQ  & 2 ($g=64$,\textit{2.28}) & 2.1e5/1.6e5  & 1.2e5/9.5e4  \\

OmniQuant  & 2 ($g=64$,\textit{2.28})  & 9.62/12.72   & 7.56/10.05    \\
\textbf{CLAQ*} & 2.12  & \textbf{7.69/9.13} & \textbf{6.30/7.94}  \\
\textbf{CLAQ*}  & 2.24 & \textbf{6.89/8.47} & \textbf{5.88/7.50}   \\     
\bottomrule
\end{tabular}
\end{table}

\begin{table}[H]
\centering
\caption{Experimental results of perplexity in Yi-34B-Base model. CLAQ models with * are fusion models employing column-level adaptive precision and outlier reservation
techniques.}
\label{table:9}
\begin{tabular}{cccc}
\toprule
\textbf{Method}      & \textbf{\# Bits}      & \textbf{\begin{tabular}[c]{@{}c@{}}Yi-34B\\ WikiText2↓/C4↓\end{tabular}} \\
\toprule
    -                 & 16                   & 24.73/29.57                                                              \\
GPTQ                 & 4                    & 32.10/40.55                                                              \\
\textbf{CLAQ}        & 4                    & \textbf{26.38/31.38}                                                              \\
\midrule
GPTQ                 & 3                    & 59.16/69.45                                                              \\
\textbf{CLAQ}        & 3                    & \textbf{56.03/67.69}                                                     \\
\textbf{CLAQ*}        & 3.12                 & \textbf{26.85/31.82}                                                     \\
\textbf{CLAQ*}        & 3.23                 & \textbf{28.89/33.24}                                                     \\
\midrule  
GPTQ                 & 2                    & 9577.46/9018.70                                                          \\
\textbf{CLAQ}        & 2                    & \textbf{199.86/247.80}                                                   \\
\textbf{CLAQ*}        & 2.12                 & \textbf{46.04/51.31}                                                     \\
\textbf{CLAQ*}        & 2.24                 & \textbf{41.80/45.90}                                                     \\
\bottomrule
\end{tabular}
\end{table}

\begin{table}[H]
\centering
\caption{Experimental results of zero-shot tasks in LLaMA-2-Base and Yi-34B-Base models.}
\resizebox{\textwidth}{!}{
    \begin{tabular}{@{}ccccccccccc@{}}
    \toprule
\textbf{Model} &\textbf{Method} & \textbf{\# Bits}  & \textbf{PIQA} & \textbf{Arc-e} &\textbf{Arc-c} & \textbf{BoolQ} & \textbf{HellaSwag}& \textbf{Winogrande} &\textbf{Avg↑} \\

\midrule
 & FP16 & 16 & 79.11 & 76.30 & 46.42 & 77.77 & 75.99 & 69.06 & 70.78 \\
 & GPTQ & 4 & 79.05 & 74.71 & 45.14 & 77.58 & 74.33 & 69.30 & 70.02 \\
\multirow{-3}{*}{\begin{tabular}[c]{@{}c@{}}LLaMA2-7B\end{tabular}} & CLAQ & 4 & 78.45 & 75.76 & 44.71 & 77.25 & 75.17 & 68.03 & \textbf{69.90} \\
\midrule
 & FP16 & 16 & 80.52 & 79.46 & 49.15 & 80.58 & 79.38 & 72.22 & 73.55 \\
 & GPTQ & 4 & 80.41 & 78.58 & 47.18 & 79.27 & 78.08 & 71.51 & 72.51 \\
\multirow{-3}{*}{\begin{tabular}[c]{@{}c@{}}LLaMA2-13B\end{tabular}} & CLAQ & 4 & 78.51 & 79.21 & 50.43 & 78.60 & 79.76 & 72.22 & \textbf{73.12} \\
\midrule
 & FP16 & 16 & 82.75 & 84.34 & 62.12 & 88.35 & 83.64 & 79.24 & 80.07 \\
 & GPTQ & 4 & 81.66 & 81.48 & 59.04 & 86.15 & 81.95 & 76.16 & 77.74 \\
\multirow{-3}{*}{\begin{tabular}[c]{@{}c@{}}Yi-34B\end{tabular}} & CLAQ & 4 & 82.64 & 83.29 & 60.15 & 88.59 & 83.26 & 77.19 & \textbf{79.19} \\
\bottomrule
\end{tabular}}
\end{table}

\begin{table}[H]
\centering
\caption{Experimental results of zero-shot accuracy in LLaMA-2-Base and Yi-34B-Base models. CLAQ models with* are fusion models employing column-level adaptive precision and outlier reservation techniques.}
\resizebox{\textwidth}{!}{
\begin{tabular}{ccccccccccc}
    \toprule
\textbf{Model} & \textbf{Method} & \textbf{\# Bits} & \textbf{PIQA} & \textbf{Arc-e} &\textbf{Arc-c} & \textbf{BoolQ} & \textbf{HellaSwag}& \textbf{Winogrande} &\textbf{Avg↑} \\
\midrule
 & FP16 & 16 & 79.11 & 75.25 & 44.62 & 75.11 & 76.20 & 70.01 & 70.38 \\
 & GPTQ & 3 & 75.52 & 64.10 & 35.58 & 66.76 & 68.31 & 63.30 & 62.26 \\
 & \textbf{CLAQ*} & 3.1 & 78.89 & 72.60 & 42.92 & 73.52 & 74.32 & 69.69 & 68.66 \\
\cmidrule(l){2-10}
 & GPTQ & 2 & 52.50 & 25.76 & 26.96 & 41.87 & 25.77 & 52.49 & 37.56 \\
\multirow{-5}{*}{\begin{tabular}[c]{@{}c@{}}LLaMA1-7B\end{tabular}} & \textbf{CLAQ*} & 2.1 & 75.73 & 68.56 & 36.43 & 70.73 & 67.05 & 67.25 & 64.29 \\
\midrule
 & FP16 & 16 & 80.14 & 77.40 & 47.70 & 77.92 & 79.09 & 72.85 & 72.52 \\
 & GPTQ & 3 & 77.69 & 73.40 & 42.75 & 68.56 & 73.14 & 66.54 & 67.01 \\
 & \textbf{CLAQ*} & 3.1 & 79.92 & 76.43 & 45.90 & 76.27 & 78.15 & 71.35 & 71.34 \\
 \cmidrule(l){2-10}
 & GPTQ & 2 & 52.18 & 25.88 & 28.41 & 41.35 & 25.67 & 49.64 & 37.19 \\
\multirow{-5}{*}{\begin{tabular}[c]{@{}c@{}}LLaMA1-13B\end{tabular}} & \textbf{CLAQ*} & 2.1 & 77.80 & 72.22 & 41.21 & 69.42 & 71.86 & 69.14 & 66.94 \\
\midrule
 & FP16 & 16 & 82.26 & 80.43 & 52.90 & 82.69 & 82.63 & 75.85 & 76.13 \\
 & GPTQ & 3 & 78.78 & 73.65 & 46.50 & 75.78 & 77.86 & 74.59 & 71.19 \\
 & \textbf{CLAQ*} & 3.1 & 81.66 & 78.96 & 51.02 & 82.11 & 81.49 & 74.66 & 74.98 \\
\cmidrule(l){2-10}
 & GPTQ & 2 & 53.16 & 27.74 & 26.79 & 43.46 & 27.10 & 48.46 & 37.79 \\
\multirow{-5}{*}{\begin{tabular}[c]{@{}c@{}}LLaMA1-30B\end{tabular}} & \textbf{CLAQ*} & 2.1 & 80.03 & 76.73 & 47.10 & 79.51 & 76.55 & 73.64 & 72.26 \\
 \midrule
 & FP16 & 16 & 82.26 & 81.36 & 55.55 & 84.83 & 84.12 & 77.27 & 77.57 \\
 & GPTQ & 3 & 80.67 & 78.07 & 51.45 & 80.15 & 80.96 & 74.35 & 74.28 \\
 & \textbf{CLAQ*} & 3.1 & 81.88 & 80.51 & 55.72 & 84.43 & 83.47 & 77.11 & 77.19 \\
\cmidrule(l){2-10}
 & GPTQ & 2 & 53.59 & 28.62 & 27.22 & 54.80 & 30.09 & 51.62 & 40.99 \\
\multirow{-5}{*}{\begin{tabular}[c]{@{}c@{}}LLaMA1-65B\end{tabular}} & \textbf{CLAQ*} & 2.1 & 80.20 & 78.96 & 50.17 & 79.14 & 79.39 & 75.30 & 73.86 \\
 \midrule
 & FP16 & 16 & 79.11 & 76.30 & 46.42 & 77.77 & 75.99 & 69.06 & 70.78 \\
 & GPTQ & 3 & 74.81 & 66.04 & 38.65 & 69.63 & 66.70 & 63.54 & 63.23 \\
 & \textbf{CLAQ*} & 3.1 & 78.62 & 74.24 & 42.83 & 75.47 & 74.33 & 69.53 & 69.17 \\
\cmidrule(l){2-10}
 & GPTQ & 2 & 52.12 & 26.52 & 27.90 & 42.45 & 26.11 & 48.07 & 37.20 \\
\multirow{-5}{*}{\begin{tabular}[c]{@{}c@{}}LLaMA2-7B\end{tabular}} & \textbf{CLAQ*} & 2.1 & 75.03 & 70.50 & 38.23 & 72.81 & 66.82 & 66.77 & 65.03 \\
 \midrule
 & FP16 & 16 & 80.52 & 79.46 & 49.15 & 80.58 & 79.38 & 72.22 & 73.55 \\
 & GPTQ & 3 & 77.97 & 73.02 & 43.86 & 74.07 & 73.88 & 68.11 & 68.49 \\
 & \textbf{CLAQ*} & 3.1 & 79.54 & 78.07 & 49.23 & 76.67 & 77.81 & 71.43 & 71.63 \\
\cmidrule(l){2-10}
 & GPTQ & 2 & 53.16 & 25.67 & 27.90 & 39.69 & 25.61 & 49.88 & 36.99 \\
\multirow{-5}{*}{\begin{tabular}[c]{@{}c@{}}LLaMA2-13B\end{tabular}} & \textbf{CLAQ*} & 2.1 & 77.53 & 74.75 & 43.43 & 76.88 & 72.79 & 70.80 & 69.36 \\
 \midrule
 & FP16 & 16 & 82.75 & 84.34 & 62.12 & 88.35 & 83.64 & 79.24 & 80.07 \\
 & GPTQ & 3 & 75.30 & 64.23 & 38.23 & 64.98 & 67.93 & 64.72 & 62.57 \\
 & \textbf{CLAQ*} & 3.1 & 83.30 & 83.33 & 58.79 & 87.86 & 82.56 & 77.35 & 78.87 \\
\cmidrule(l){2-10}
 & GPTQ & 2 & 51.90 & 25.04 & 27.30 & 40.52 & 25.97 & 48.07 & 36.47 \\
\multirow{-5}{*}{\begin{tabular}[c]{@{}c@{}}Yi-34B\end{tabular}} & \textbf{CLAQ*} & 2.1 & 81.07 & 80.43 & 54.27 & 79.36 & 77.49 & 74.90 & 74.59 \\
\bottomrule
\end{tabular}}
\end{table}




\section{Experiment details}
Our experiments are conducted based on a branching off from the GPTQ\cite{frantar2022gptq} implementation. The quantization group size is set to the entire column. The calibration data consists of 128 random 2048 token segments from the C4 dataset. The K-Means clustering is calculated with the scikit-learn-intelex CPU version.

The fusion models with AP+OR utilize two settings: (1) 2.12/3.12bit: 0.05 bit increment in adaptive precision with 2\&4 bits, and 0.07 bit of full-precision outliers kept with Setting 2 in Appendix C. (2) 2.24/3.23bit: 0.1 bit increment in adaptive precision with 2\&4 bits, and 0.13 bit of full-precision outliers kept with Setting 2 in Appendix C. The outlier standard is set to $S=13$ in all experiments. Based on our observations in Appendix A, we fixed the column proportion of outliers to be retained in our experiments. We always preserve more outliers in the most sensitive top 10\% of columns, while retaining fewer outliers in the remaining 90\% of columns. The code for reproducing the results in Table \ref{table:1}, Table \ref{table:2} and other tables in Appendix E is contained in supplemental materials. Check the ReadMe file for more information.

\section{Heuristic adaptive precision search algorithm}
Our adaptive precision approach, detailed in Section 3.3, combines two precision levels. However, when dealing with expanded adaptive precision search landscapes, such as 2.5 bits instead of 2.1 bits, straightforward methodologies may not consistently discover the most advantageous configurations. To address this complexity, we propose a heuristic-based adaptive precision search algorithm. This algorithm, inspired by the HAWQ v2 \cite{dong2020hawq} method, initially ranks weight matrices by their outlier ratios, ensuring matrices with higher outlier concentrations receive higher precision. This step transforms the search space into a combinatorial problem.

Next, we discretize the precision levels for each column to a defined set, thereby confining the search within a manageable subset of all possible combinations under the model size constraint. We traverse this reduced search space, exploring all feasible precision arrangements, and establish a precision scoring system to facilitate the selection of the most suitable configuration. In essence, this heuristic-driven adaptive precision search process defines discrete precision options for each column, enumerates all feasible combinations within the narrowed search space, and selects the combination with the highest precision score. This systematic approach optimizes the use of precision levels across matrices, thereby enhancing the overall quantization efficiency under practical constraints.

Specifically, in the case of an adaptive precision search involving 2-bit and 3-bit quantization levels, we categorize each matrix's potential precision into three classes: 2-bit, a combination of 2\&3-bit, and a combination of 2\&4-bit. We allocate a higher precision score $PS_4$ to the 4-bit columns in 2\&4-bit combination and a score $PS_3$ for 3-bit parameters in the 2\&3-bit matrices. The total score $PS_{total}$ can be formulated as:

\begin{gather}
    P=\left \{ p_3,p_4 \right \}  \quad M=\left \{ M_3,M_4 \right \}  \\
PS_{total}(P,L) = OR_4\times PS_4 \times p_4 \times M_4+OR_3\times PS_3 \times p_3 \times M_3 \\
P,L = \mathop{\arg\max}_{q}  \enspace PS_{total}
\end{gather}

Where $P$ is the percentage of high-precision columns in weight matrices, $M$ is the numbers of high-precision matrices. $OR$ denotes the outlier ratio of high-precision matrices. $PS_3,PS_4$ are precision scores defined to calculate the total precision score in the model. We set $PS_3=3,PS_4=4$ for experiments in Table \ref{table:12}. The precision score for 2-bit columns are 0. This heuristic search strategy realizes adaptive precision quantization by exhaustively enumerating all feasible combinations of precision allocations, subsequently computing a search criterion based on the weighted parameter precision scores, which are derived from the outlier ratios. The goal of adaptive precision search process is to find the combination with the highest precision score. Following this adaptive precision search strategy, in scenarios where the incremental bit-width is modest (i.e., 2.1 or 2.2 bits), the search results favor a configuration utilizing the maximum number of 2\&4-bit matrices, which is similar to our previously described approach. In the case of a 2.5-bit search, 19 matrices allocated as 2\&4-bit with 10\% 4-bit columns, 205 2\&3-bit matrices are allocated with 52.6\% 3-bit columns. Nevertheless, the model consistently minimizes the proportion of parameters quantized solely at 2 bits. The results of the heuristic algorithm based adaptive precision quantization are detailed in Table \ref{table:12}, evidencing that heuristic-based searching can indeed refine model performance.

\begin{table}[H]
\centering
\caption{Results of heuristic adaptive precision search at 2.5-bit.}
\label{table:12}
\resizebox{\linewidth}{!}{
\begin{tabular}{lcccccccccc}
\toprule
\textbf{Model} & \textbf{\# Bits} & \textbf{WikiText2↓} & \textbf{C4↓}  & \textbf{PIQA} & \textbf{Arc-e} &\textbf{Arc-c} & \textbf{BoolQ} & \textbf{HellaSwag}& \textbf{Winogrande} &\textbf{Avg↑} \\
\midrule
LLaMA1-7B       & 16               & 5.63     & 7.08   & 79.11   & 75.25  & 44.62   & 75.11        & 76.20  & 70.01  & 70.38   \\
CLAQ  & 3   & 6.47     & 7.87  & 78.23    & 72.05   & 41.12  & 72.93  & 72.05            & 67.71   & 67.35      \\
CLAQ     & 2     & 27.64     &24.37 & 62.89    & 31.43   & 25.51           & 61.52       & 28.69   & 51.85   & 43.65   \\
\midrule

 \textbf{+AP(ours)}   & 2.5    & 8.43       & 9.78   & 75.63  & 68.10 & 35.24  & 66.39   & 63.82   & 64.40        & 62.26    \\
  \textbf{+AP(Heuristic AP search)}   & 2.5   & \textbf{7.88}  & \textbf{9.40}    & 75.95         & 68.73   & 36.01   & 72.39   & 65.98   & 64.72     & \textbf{63.96}   \\
\bottomrule
\end{tabular}}
\end{table}

However, further analyses indicate that the stability of our heuristic adaptive precision search strategy is somewhat inconsistent, with results being heavily influenced by the precision score assignment. This highlights a key insight: developing an end-to-end, effective, and interpretable adaptive precision search strategy that consistently improves performance under varying conditions remains a significant challenge for future research. The pursuit of a robust and explainable algorithm capable of systematically navigating the extensive search-space of adaptive precision configurations, optimizing both model performance and computational efficiency, is a critical direction for advancing the field of model quantization.

\section{Ablation study on calibration data}
We investigated the impact of calibration data on experimental outcomes, aligning with the data selection used in GPTQ \cite{frantar2022gptq}. Calibration data consists of 128 random 2048 token segments from the C4 \cite{raffel2020c4} dataset. We found the non-negligible effect of varying calibration datasets, emphasizing that tailoring the calibration data to specific datasets positively influences the perplexity (PPL) evaluation for those datasets. This highlights the importance of a calibrated and dataset-specific approach in enhancing quantization performance.

\begin{table}[H]
\centering
\caption{Experimental results of CLAQ models calibrated on C4 \cite{raffel2020c4} and Wikitext2 \cite{merity2016wiki2}.}
\begin{tabular}{ccccc}
\toprule
\textbf{Model} & \textbf{\# Bits} & \textbf{Calibration} & \textbf{WikiText2↓} & \textbf{C4↓} \\
\midrule
LLaMA1-7B & 16 & -- & 5.63 & 7.08 \\
\midrule
CLAQ & 4 & on wiki & 5.72 & 7.23 \\
CLAQ & 4 & on c4 & 5.78 & 7.21 \\
\midrule
CLAQ & 3 & on wiki & 6.04 & 7.89 \\
CLAQ & 3 & on c4 & 6.47 & 7.87 \\
\midrule
CLAQ & 2 & on wiki & 32.35 & 44.95 \\
CLAQ & 2 & on c4 & 27.64 & 24.37 \\
\bottomrule
\end{tabular}
\end{table}

\section{Broader Impact}

Our proposed LLM compression approach has the potential to enhance the compression ratio of mainstream open-source LLMs by up to 8x, with nearly lossless compression enhancement reaching up to 4x. This expands the applicability of LLMs. However, it is beyond our scope to determine whether the compressed models maintain their intended behavior post-compression; we lack the capability to monitor the exact usage scenarios of our compressed models or ascertain whether they might be repurposed for unintended applications.

\end{document}